\pdfoutput=1

\documentclass[11pt]{article}

\usepackage{acl}

\usepackage{times}
\usepackage{latexsym}

\usepackage[T1]{fontenc}

\usepackage[utf8]{inputenc}
\usepackage[main=english,russian,ukrainian,spanish]{babel}

\usepackage{microtype}

\usepackage{inconsolata}
\usepackage{booktabs}
\usepackage{graphicx}
\usepackage{color, colortbl}
\usepackage{enumitem}
\usepackage{caption}
\usepackage{adjustbox}
\usepackage{amssymb}
\usepackage{utfsym}
\usepackage{mathtools}
\usepackage{amsmath}
\usepackage{subcaption}
\usepackage{enumitem}

\usepackage{xcolor}

\definecolor{Gray}{gray}{0.9}

\title{MultiParaDetox: Extending Text Detoxification \\ with Parallel Data to New Languages}

\author{
\textbf{Daryna Dementieva\textsuperscript{1}}, \textbf{Nikolay Babakov}\textsuperscript{2}, \textbf{and}
\textbf{Alexander Panchenko\textsuperscript{3,4}} \\
\textsuperscript{1}Technical University of Munich \\ \textsuperscript{2}Centro Singular de Investigación en Tecnoloxías Intelixentes (CiTIUS), \\ Universidade de Santiago de Compostela \\
\textsuperscript{3}Skolkovo Institute of Science and Technology, \textsuperscript{4}Artificial Intelligence Research Institute\\ 
\href{mailto:daryna.dementieva@tum.de}{\texttt{\small daryna.dementieva@tum.de}},
\href{mailto:nikolay.babakov@usc.es}{\texttt{\small nikolay.babakov@usc.es}},
\href{mailto:a.panchenko@skol.tech}{\texttt{\small a.panchenko@skol.tech}}
}

\begin{document}
\maketitle
\begin{abstract}
Text detoxification is a textual style transfer (TST) task where a text is paraphrased from a toxic surface form, e.g. featuring rude words, to the neutral register. Recently, text detoxification methods found their applications in various task such as detoxification of Large Language Models (LLMs) \cite{leong-etal-2023-self,DBLP:journals/corr/abs-2308-05596,DBLP:journals/corr/abs-2308-08295} and toxic speech combating in social networks \cite{DBLP:journals/corr/abs-2302-09270,mun-etal-2023-beyond,DBLP:journals/corr/abs-2310-13985}. 
All these applications are extremely important to ensure safe communication in modern digital worlds. 
However, the previous approaches for parallel text detoxification corpora collection---ParaDetox \cite{DBLP:conf/acl/LogachevaDUMDKS22} and APPADIA \cite{DBLP:conf/coling/AtwellHA22}---were explored only in monolingual setup. In this work, we aim to extend ParaDetox pipeline to multiple languages presenting \textbf{MultiParaDetox} to automate parallel detoxification corpus collection for potentially any language. Then, we experiment with different text detoxification models---from unsupervised baselines to LLMs and fine-tuned models on the presented parallel corpora---showing the great benefit of parallel corpus presence to obtain state-of-the-art text detoxification models for any language. 

\textcolor{red}{\textit{Warning: This paper contains rude texts that only serve as illustrative examples.}}
\end{abstract}

\section{Introduction}

We formulate text detoxification task as stated in \cite{DBLP:journals/mti/DementievaMLDKS21} so the objective is to paraphrase a toxic text to a text that: (i) has neutral style (register); (ii) saves the meaningful content as much as possible; (iii) is fluent at least at the same level as the input text. Before, many unsupervised approaches for text detoxification were presented \cite{nogueira-dos-santos-etal-2018-fighting,DBLP:conf/emnlp/DaleVDLKSP21,floto-etal-2023-diffudetox} addressing the task based only on available toxic or hate speech classification corpora which are most commonly non-parallel. However, in ParaDetox \cite{DBLP:conf/acl/LogachevaDUMDKS22} and APPADIA \cite{DBLP:conf/coling/AtwellHA22} the benefit of parallel corpus for text detoxification was illustrated---the \textit{seq2seq} models like BART \cite{lewis-etal-2020-bart} and T5 \cite{DBLP:journals/jmlr/RaffelSRLNMZLL20} fine-tuned on the presented corpora outperformed previous unsupervised baselines in both manual and automated evaluations. 

\begin{table}[t!]
    \centering
    \footnotesize
    \begin{tabular}{p{1.70cm}|p{5.3cm}}
    \toprule
        
        \multicolumn{2}{c}{\textbf{Russian}} \\
        \midrule 
        Original \newline \newline Detox & \foreignlanguage{russian}{Тебя это е**ть не должно, п***рюга} \newline \textcolor{gray}{\scriptsize{\textit{You shouldn't give a f**k, f**got}}} \newline\foreignlanguage{russian}{Тебя это волновать не должно} \newline \textcolor{gray}{\scriptsize{\textit{You don't have to worry about that}}} \\
        \midrule 

        \multicolumn{2}{c}{\textbf{Ukrainian}} \\
        \midrule 
        Original \newline \newline \newline Detox & \foreignlanguage{ukrainian}{С**а як же мене всі бісять б**ть н**уй} \newline \textcolor{gray}{\scriptsize{\textit{F**k, everyone pisses me the f**k off}}} \newline \foreignlanguage{ukrainian}{як же мене всі бісять} \newline \textcolor{gray}{\scriptsize{\textit{I'm so irritated by everyone}}} \\
        \midrule 

        \multicolumn{2}{c}{\textbf{Spanish}} \\
        \midrule 
        Original \newline \newline Detox & \foreignlanguage{spanish}{Este país se va a la m**rda} \newline \textcolor{gray}{\scriptsize{\textit{This country is going to s**t}}} \newline \foreignlanguage{spanish}{Cosas van muy mal en este país} \newline \textcolor{gray}{\scriptsize{\textit{Things are going very badly in this country}}} \\

        \bottomrule
    \end{tabular}
    \caption{Text detoxification parallel pairs examples from Russian, Ukrainian, and Spanish \textbf{ParaDetox} datasets.}
    \label{tab:intro_examples}
\end{table}

\begin{figure*}[th!]
    \centering
    \includegraphics[width=\textwidth]{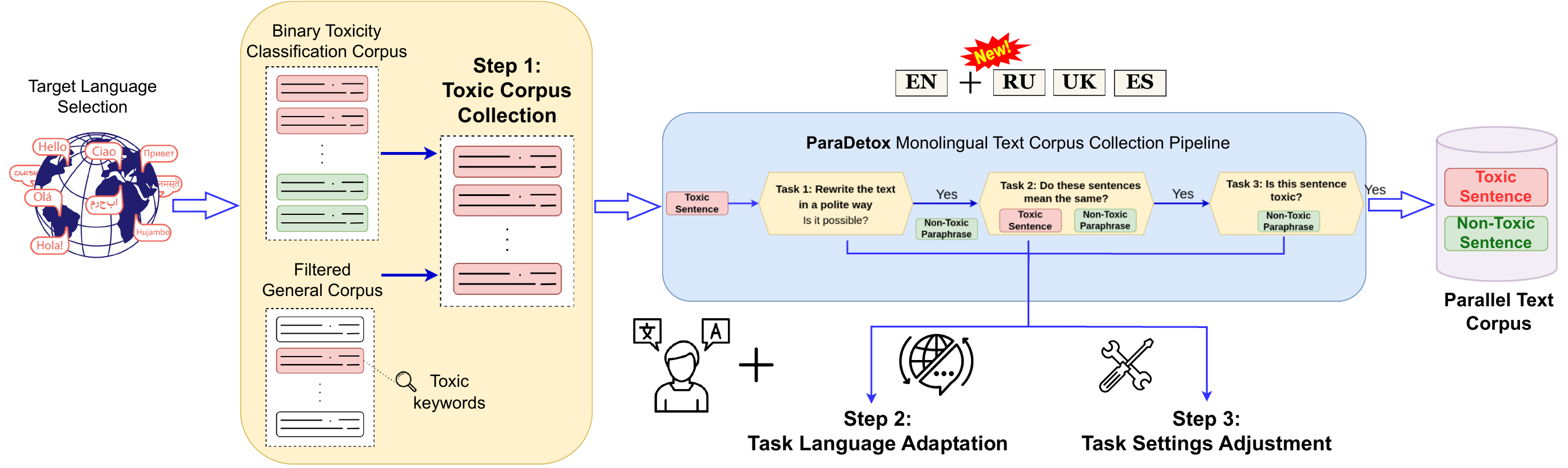}
    \caption{\textbf{MultiParaDetox} pipeline for parallel corpus collection using crowsourcing: \textbf{Step 1}: \textit{Toxic Corpus Collection} - texts can obtained either from available for the target language binary classification (non-parallel) corpus or by keywords search in some general corpus; \textbf{Step 2:} \textit{Task Language Adaptation} to the target language with translation system and a cross-check by native speakers; \textbf{Step 3:} \textit{Tasks Settings Adjustment} by configuring annotators language requirements and quality control tasks.}
    \label{fig:MultiParaDetox}
\end{figure*}

While the parallel detoxification corpora are already available together with their collection pipelines, they were only presented for English language. However, we strongly support the idea of such corpus availability for any language would lead to fair and safe LMs development equally for all languages \cite{DBLP:journals/corr/abs-2212-04960}. In this work, we aim to extend ParaDetox collection pipeline to a multilingual format confirming the \textit{hypothesis} that it can be used to collect parallel text detoxification data for any language\footnote{In our study we use crowdsourcing platforms: they have wide, yet limited support of languages. In principle, our pipeline shall be usable for spoken languages with available text corpora (preferably in form of user-generated comments).}. 
Thus, the contributions of this work are as following:
\begin{itemize}
    \item We present MultiParaDetox: a pipeline for extension of text detoxification corpus collection procedure to new languages;
    \item We showcase the pipeline collecting new parallel datasets for three new languages---Spanish (from Romance branch of Indo-European language family), Russian, and Ukrainian (from East Slavic branch);
    \item We present the first of its kind evaluation study of unsupervised baselines, LLMs, and fine-tuned supervised models for these three languages for the text detoxification task affirming the advantages of parallel corpora. 
\end{itemize}

All the introduced data and models are available for public usage online.\footnote{\href{https://huggingface.co/textdetox}{https://huggingface.co/textdetox}}

\section{Related Work}
\paragraph{Text Style Transfer with Parallel Data}
While the tasks of text style transfer were explored for diverse domains (sentiment, authors styles, formality, toxicity), these problems are addressed in the majority of cases only with non-parallel text classification corpora.
To this date, only a few parallel corpora for text style transfer were presented: (i) Bible corpus \cite{carlson2018evaluating} which was obtained historically due to many reissues of the text; (ii) GYAFC \cite{rao-tetreault-2018-dear} which was collected via crowdsourcing but verified manually by the authors of the work; (iii) APPADIA \cite{DBLP:conf/coling/AtwellHA22} which was annotated by expert sociolinguists; (iv) ParaDetox \cite{DBLP:conf/acl/LogachevaDUMDKS22} which was fully collected and verified by crowdsourcing. 

\paragraph{Text Detoxification} As ParaDetox and APPADIA datasets have appeared recently, the vast attention in the text detoxification field has been paid to unsupervised methods. In \cite{nogueira-dos-santos-etal-2018-fighting}, the basic encoder-decoder was extended with a collaborative classifier and a set of specialized loss functions for detoxification. Then, the power of Masked Language Modelling (MLM) were utilized in CondBERT and ParaGedi models \cite{DBLP:conf/emnlp/DaleVDLKSP21}. These unsupervised baselines were improved with the mixture of experts and anti-experts concept in MaRCo \cite{DBLP:conf/acl/HallinanL0S23}. However, the seq2seq models from ParaDetox and APPADIA works showed so far more promising text detoxification results than those based on non-parallel corpora, such as those mentioned above.

\paragraph{Text Style Transfer in Multilingual and Cross-lingual Setups} Also, several works are already dedicated to the extension of text style transfer methods to new languages. For sentiment, in \cite{mukherjee-etal-2023-low}, English dataset was extended to Bangla with manual annotation. X-FORMAL dataset \cite{briakou-etal-2021-ola} was introduced as the extension of GYAFC for three new languages and was obtained via automated translation. For text detoxification task, the cross-lingual setup was explored in \cite{dementieva-etal-2023-exploring} attempting to transfer knowledge from English to a low-resource language. While several approaches showed compatible results, they are still inferior in quality to methods fine-tuned on parallel data.

\section{MultiParaDetox Pipeline}

\begin{table*}[ht!]
    \centering

\resizebox{1.0\textwidth}{!}{

    \begin{tabular}{c|c|c|c|c|c|c|c|c}
        \toprule
        \shortstack{\bf Target \\ \bf  Language} & \shortstack{ \bf Input \\ \bf  Samples} & \shortstack{\bf Filtered \\\bf  Non-detox. \\ \bf Samples} & \shortstack{\bf Unique \\ \bf  Inputs \\ \bf  Paraphrases} & \shortstack{\bf Paraphrases \\ \bf per Input} & \shortstack{ \bf Paraphrases \\ \bf Total} & \shortstack{\bf Length in \\\bf Tokens of \\ \bf Toxic/neutral} & \shortstack{\bf Total \\ \bf Costs} & \shortstack{\bf Cost per \\ \bf Unique \\ \bf Sample} \\ \midrule
         Russian & 30\,000 & 65\% & 8\,500 & 1.83 & 11\,200 & 10.1 / 9.7 & \$880 & \$0.11 \\
        \midrule
         Ukrainian & 2\,700 & 20\% & 2\,122 & 2.19 & 4\,661 & 12.5 / 10.8 & \$849 & \$0.18 \\
        \midrule
         Spanish & 720 & 54\% & 337 & 1.67 & 565 & 11.7 / 9.6 & \$278 & \$0.40 \\
        
        \bottomrule
    \end{tabular}

}
    
    \caption{Statistics of new ParaDetox data: the crowdsourcing steps and final datasets.}
    \label{tab:generated_para_info}
\end{table*}

We adapt ParaDetox \cite{DBLP:conf/acl/LogachevaDUMDKS22} collection pipeline as it was designed to automate the data collection as well as verification with crowdsourcing. The pipeline consists of three tasks: 
\begin{description}
\item[\textbf{Task 1: Rewrite text in a polite way}] Annotators need to provide the detoxified paraphrase of the text so it becomes non-toxic and the main content is saved or to skip paraphrasing if the text is not possible to rewrite in non-toxic way; 
\item[\textbf{Task 2: Do these sentences mean the same?}] Check if the content is indeed the same between the original toxic text and its potential non-toxic paraphrase; 
\item[\textbf{Task 3: Is this text offensive?}] Verification of the provided paraphrase if it is indeed non-toxic.
\end{description}

We extend this pipeline to \textbf{MultiParaDetox} supporting any new language (see Figure~\ref{fig:MultiParaDetox}):

\paragraph{Step 1: Toxic Corpus Preparation} Firstly, we need to prepare toxic samples that will serve as input to the ParaDetox pipeline. In the annotation, we focus only on \textit{explicit} toxicity types \cite{van-aken-etal-2018-challenges}. (i) If there already exists binary toxicity classification dataset, then it is enough to select from it toxic or hate part preferably with labels like \textit{``toxic''}, \textit{``offensive''}, \textit{``obscene''}; (ii) If there is not such dataset, then samples with explicit toxicity can be selected by finding toxic keywords substrings in the texts. As we want sentences to have meaningful content, only sentences with less then $1/2$ of toxic keywords fraction should be chosen.

\paragraph{Step 2: Tasks Language Adaptation} Then, the ParaDetox tasks needed to be adapted for the target language. This can be done with combination of automated translation followed by language native speakers texts proofreading.

\paragraph{Step 3: Tasks Settings Adjustment} Finally, for the crowdsourcing tasks, the language and country settings should be chosen accordingly to the target language. For the quality control, we follow the procedure described in ParaDetox utilising training, exam, and control tasks. We claim that these tasks can be also translated from the original ones with slight edits by native speakers according to special cases of toxicity for the language.

\section{Collection of New Parallel Datasets with MultiParaDetox}
We applied the described above pipeline to obtain new parallel datasets for three languages---Russian, Ukrainian, and Spanish. The choice of language was done based on the availability of native speakers of these languages. The data collection was done via Toloka platform.\footnote{\href{https://toloka.ai}{https://toloka.ai}} For translations, we used DeepL\footnote{\href{https://www.deepl.com/translator}{https://www.deepl.com/translator}} API (tasks texts are presented in Appendix~\ref{sec:app_multi_paradetox_tasks}). We accepted to the annotation only workers who proved the corresponding language fluency with a test. The general information about the datasets is presented in Table~\ref{tab:generated_para_info} with example samples in Appendix~\ref{sec:app_multi_paradetox_examples}.

As Russian toxicity classification datasets were available for Russian language, we selected toxic sentences from Russian Language Toxic Comments competitions~\cite{tox_ru_comments_1,tox_ru_comments_2}. 
%
In case of the Ukrainian language, there was no binary toxicity classification corpus available. We filtered from Ukrainian Tweets Corpus~\cite{bobrovnyk2019automated} the explicitly toxic samples that contain obscene lexicon from the predefined list~\cite{tox_uk_lexicon}.
%
For Spanish language, we selected samples for annotation from two datasets: hate speech detection one~\cite{DBLP:journals/sensors/Pereira-Kohatsu19} as well as filtered by keywords Spanish Tweets corpus~\cite{perez-etal-2022-robertuito}.

We collected the data for each language depending on the available input data and resources. The nature of the original data influenced the process. Thus, the lowest ratio of non-detoxified sample filtering was observed for Ukrainian language. For Spanish, this ratio is higher as the input data labels were more from the hate speech domain. Nevertheless, for each language it was possible to collect at least several hundreds of pairs with 1-3 paraphrases per each toxic input.

\paragraph{Data Quality Verification}
To verify the quality of all collected data, we randomly selected $100$ pairs per language and asked 3 annotators---native-speakers for each language with the expertise in the topic---to label if the pair meets the requirements of the task or not. For all languages, the amount of inappropriate pairs was $<10\%$. The inner-annotator agreement was estimated with Krippendorff's $\alpha$: for Russian it equals to 0.85, for Ukrainian it equals to 0.90, and for Spanish it equals to 0.67.

\section{Text Detoxification Experiments}
To enhance the validation of the data collected using MultiParaDetox, we conduct text detoxification experiments, comparing baselines with fine-tuned models with the newly obtained data.

\subsection{Text Detoxification Systems}

\paragraph{Duplicate} Simple copy-paste of the toxic input to the output without any change. This baseline has the highest SIM score by definition. 

\paragraph{Delete} Elimination of obscene substrings from a manually constructed dictionary of rude words. Existing lexicons are used for Russian~\cite{DBLP:journals/mti/DementievaMLDKS21}, Ukrainian~\cite{tox_uk_lexicon}, and Spanish~\cite{tox_es_lexicon}.

\paragraph{condBERT} We adapted one of the MLM-based unsupervised methods from~\cite{DBLP:conf/emnlp/DaleVDLKSP21}. We used mBERT~\cite{devlin-etal-2019-bert} as a base model. The model runs MLM to generate list of substitutes selecting non-toxic ones.

\paragraph{LLM Prompting} Firstly, we experimented with several multilingual models---MT0-large~\cite{DBLP:conf/acl/MuennighoffWSRB23}, BloomZ-7b~\cite{DBLP:conf/acl/MuennighoffWSRB23}, and LLaMa-7b~\cite{DBLP:journals/corr/abs-2307-09288}---to select the most promising one for the task (see the results in Appendix~\ref{sec:app_llm_selection}). In the end, we proceed with LLaMa
 in zero-shot setup with the corresponding for each language prompt: 
 
\textsf{\small Rewrite the following text in a more polite but natural form, maintaining its original meaning (no comments, just rewritten text) \{text\}}.

\paragraph{Fine-tuned LM on Translated Data} We also tried to obtain synthetic parallel corpa by translating English ParaDetox~\cite{DBLP:conf/acl/LogachevaDUMDKS22} to our target languages. We utilized mBART model~\cite{DBLP:journals/tacl/LiuGGLEGLZ20}\footnote{\href{https://huggingface.co/facebook/mbart-large-50-many-to-many-mmt}{https://huggingface.co/facebook/mbart-large-50-many-to-many-mmt}} for the translation step.

\paragraph{Fine-tuned LM on ParaDetox} Finally, we fine-tuned text generation models on the presented data. We fine-tuned mBART~\cite{DBLP:journals/tacl/LiuGGLEGLZ20}\footnote{\href{https://huggingface.co/facebook/mbart-large-50}{https://huggingface.co/facebook/mbart-large-50}} in both monolingual and multilingual setups.

\subsection{Evaluation Setups}
We follow the automated evaluation setup used in~\cite{DBLP:conf/acl/LogachevaDUMDKS22} adapting it to our target languages. In this setup, three following components are measured:

\paragraph{Style Transfer Accuracy (STA)} Toxicity classification result from the classifiers: for Russian~\cite{DBLP:journals/mti/DementievaMLDKS21}, Ukrainian (we trained our own classifier based on the additionally collected data with Task 3), Spanish~\cite{DBLP:journals/corr/abs-2004-06465}.

\paragraph{Content Similarity (SIM)} Cosine similarity between LaBSE embeddings~\cite{DBLP:conf/acl/FengYCA022} of a toxic input and a model's output.

\paragraph{Fluency (FL)} Perplexity score of the output from mGPT model~\cite{10.1162/tacl_a_00633} compared to the score of the input--the acceptable output should be no less fluent as input.

The three components  are subsequently combined into the final \textbf{Joint (J)} metric used for the final ranking of approaches. Given an input toxic text $x_i$ and its output detoxified version $y_i$, for a test set of $n$ samples:
\begin{center}
    $\textbf{J} = \frac{1}{n}\sum\limits_{i=1}^{n}\textbf{STA}(y_i) \cdot \textbf{SIM}(x_i,y_i) \cdot \textbf{FL}(y_i)$,
\end{center}
where \textbf{STA}($y_i$), \textbf{SIM}($x_i,y_i$), \textbf{FL}($y_i$) $\in \{0, 1\}$ for each text detoxification output sample $y_i$.

\section{Results}
The results of the systems evaluation are presented in Table~\ref{tab:new_languages_results}. Additionally, we provide the examples of models outputs in Appendix~\ref{sec:app_detoxification_results}.

Delete methods reaches the highest content similarity as it was designed to modify the original sentence slightly. However, it does not filter all toxic language and gains the lowest STA scores. The condBERT method fails to make substitutions with correct words and obtains not good enough fluency scores. LLaMa achieves very high STA scores concurrently with the lowest SIM scores. The model can hallucinate and even generate text not in a target language as can be observed from the examples. 

The models fine-tuned on the translated datasets fail for each language in STA scores. 
The rationale lies in the diversity of toxic phrases across languages. For instance, Russian and Ukrainian, being morphologically rich languages, encompass a multitude of toxic expressions that cannot be directly translated from English. Moreover, there exists a strong correlation between language and culture, manifesting in specific discussion topics and expressions unique to each language's online informational space.

Finally, the models fine-tuned on the proposed data never fail in any of the evaluation parameters and outperform unsupervised baselines based on J score with a high gap. This attests to the reliability of our data and necessity of parallel text detoxification corpora in acquiring state-of-the-art text detoxification models. For Spanish, a slight drop of the results can be cased by significantly lower amount of the training data. Even in this case, the model shows promising results while other models still did not produce qualitative results (LLaMa got high STA scores but the content output text was just random).

In the end, we also presented the results for multilingual text detoxification model fine-tuned for all three languages. The obtained results on par with monolingual models confirm the possibility to obtain single multilingual model for the multilingual text detoxification task.

\begin{table}[th!]
\footnotesize
\centering

\begin{tabular}{p{2.9cm}|c|c|c|c}
\toprule

& \textbf{STA} & \textbf{SIM} & \textbf{FL} & \textbf{J} \\ \hline
\multicolumn{5}{c}{\textbf{Russian}} \\ \hline
\rowcolor{gray!20} Human references & 0.858 & 0.720 & 0.783 & 0.484  \\
\rowcolor{gray!20} Duplicate & 0.244 & 1.000 & 1.000 & 0.247 \\ 
\hline
Delete & 0.568 & 0.891 & \textbf{0.856} & 0.410 \\
condBERT & 0.585 & 0.872 & \textcolor{gray}{0.685} & 0.349 \\
LLaMa & \textbf{0.896} & \textcolor{gray}{0.285} & 0.763 & \textcolor{gray}{0.195} \\
mBART-Translated & \textcolor{gray}{0.452} & \textbf{0.893} & 0.826 & 0.333 \\
\hline
\rowcolor{green!30} mBART-RuParaDetox & 0.772 & 0.750 & 0.781 & \textbf{0.492} \\
\hline \hline

\multicolumn{5}{c}{\textbf{Ukrainian}} \\ \hline
\rowcolor{gray!20} Human references & 0.872 & 0.897 & 0.669 & 0.523  \\
\rowcolor{gray!20} Duplicate & 0.053 & 1.000 & 1.000 & 0.053 \\ 
\hline
Delete & 0.872 & \textbf{0.944} & 0.163 & 0.134 \\
condBERT & 0.747 & 0.869 & \textcolor{gray}{0.147} & \textcolor{gray}{0.095} \\
LLaMa & \textbf{0.900} & \textcolor{gray}{0.349} & 0.669 & 0.210 \\
mBART-Translated & \textcolor{gray}{0.506} & 0.900 & 0.678 & 0.309 \\
\hline
\rowcolor{green!30} mBART-UkParaDetox & 0.759 & 0.929 & \textbf{0.725} & \textbf{0.511} \\
\hline \hline

\multicolumn{5}{c}{\textbf{Spanish}} \\ \hline
\rowcolor{gray!20} Human references & 0.653 & 0.843 & 0.407 & 0.224  \\
\rowcolor{gray!20} Duplicate & 0.195 & 1.000 & 1.000 & 0.195 \\ 
\hline
Delete & 0.415 & \textbf{0.955} & 0.305 & 0.121 \\
condBERT & 0.525 & 0.884 & \textcolor{gray}{0.161} & \textcolor{gray}{0.075} \\
LLaMa & \textbf{0.949} & \textcolor{gray}{0.284} & \textbf{1.000} & \textbf{0.269}\\
mBART-Translated & \textcolor{gray}{0.407} & 0.861 & 0.619 & 0.217 \\
\hline
\rowcolor{green!30} mBART-EsParaDetox & 0.576 & 0.858 & 0.483 & 0.239 \\
\hline \hline
\multicolumn{5}{c}{\textbf{All languages: English, Russian, Ukrainian, Spanish}} \\ \hline
mBART-MParaDetox & 0.675 & 0.958 & 0.690 & 0.456\\

\bottomrule
\end{tabular}
\caption{Text detoxification results. Within methods comparison, \textbf{bold} numbers denote the best results in a column, \textcolor{gray}{gray} -- the lowest.} 
\label{tab:new_languages_results}
\end{table}

\section{Conclusion}
We presented \textbf{MultiParaDetox}---the extension of ParaDetox pipeline for parallel data collection for the text detoxification task to new languages. The target language corpus collection can be prepared only with three steps: provision of input toxic corpus, crowdsourcing tasks language adaptation, and corresponding settings adjustments. We tested our proposed pipeline extension on three new languages---Russian, Ukrainian, and Spanish---collecting corresponding new corpora. 

The quality of the data was verified manually by native speakers. Finally, the data efficacy was confirmed with text detoxification systems comparison where the models fine-tuned on our data outperformed unsupervised baselines and zero-shot-prompted LLMs.

\section*{Limitations}
Firstly, we would like to emphasize that in our text detoxification task definition and data we purposely include only \textit{explicit} types of toxicity. More specifically, one may consider the task studied in this paper as paraphrase from the rude to neutral style. The task of addressing \textit{implicit} toxicity is more challanging \cite{wiegand-etal-2023-euphemistic} and may require different other forms of its post-processing \cite{mun-etal-2023-beyond}. While a rude text can be paraphrased to a neural form if its message is inherently non-toxic, implicitly toxic text carrying inherently toxic message hardly can be paraphrased without the change of this original toxic meaning. To collect parallel datasets for new toxicity types, i.e. sarcasm, racism, more sophisticated definition of the text detoxification task should be designed. 

Additionally, the datasets resulting from our data collection experiments exhibited an uneven distribution of sample ratios. That happened due to natural sequential progress of experiments and available resources for each step. We openly share the tasks instruction for each language so the research community can as well contribute to the data collection. Also, the further research direction might be to explore the minimal necessary amount of parallel data to fine-tune a solid text detoxification model.

While we presented the experiment to obtain one multilingual text detoxification model, the task of cross-lingual knowledge transfer between languages still has a room for improvement. Before, there was already preliminary experiments for cross-lingual text detoxification transfer~\cite{dementieva-etal-2023-exploring}. However, there is still a possibility to extension to more languages. Another side of this questions is to explore if the transfer between languages from neighbouring language families can help to improve the performance.

\section*{Ethical Considerations}
We explore the task of text detoxification only for the positive impact side of the textual communication. Thus, such systems can be potentially used in automated dialogue systems \cite{DBLP:journals/corr/abs-2302-09270}, preprocess training data~\cite{DBLP:journals/corr/abs-2308-08295}, and more niche toxicity tackling as, for instance, misogyny~\cite{DBLP:journals/corr/abs-2311-09443}. The reverse process, toxificiation of the texts, can be done simply by adding some obscene lexicon to the texts and then easily can be addressed with our models.

During crowdsourcing process, we established the most fair to our understanding payment to annotators: Task 1 -- 0.15\$ per page, Task 2 -- 0.12\$ per page, Task 3 -- 0.10\$ per page. The data were collected in several dozens of iterations and each iteration was of several hundreds of pages which resulted to the enough amount of tasks to be completed by annotators.


\bibliography{anthology,custom}

\onecolumn
\appendix

\section{MultiParaDetox Crowdsourcing Tasks and Instructions}
\label{sec:app_multi_paradetox_tasks}
Here, we list the texts of crowdsourcing task titles and instructions in their original form used to collect MultiParaDetox correspondingly for each languages: (i) Russian (Section~\ref{sec:app_multi_paradetox_tasks_russian}); (ii) Ukrainian (Section~\ref{sec:app_multi_paradetox_tasks_ukrainian}); (iii) Spanish (Section~\ref{sec:app_multi_paradetox_tasks_spanish}).

\subsection{Russian}
\label{sec:app_multi_paradetox_tasks_russian}

\textbf{Task 1:\foreignlanguage{russian}{Перепишите текст в вежливом стиле}}

\textit{Instructions}

\foreignlanguage{russian}{Вам будет показан текст, который, возможно, содержит оскорбления или грубые выражения. Вам требуется переписать его так, чтобы сохранить содержание и избавиться от оскорблений. Если текст не оскорбительный и не грубый, выберите опцию "Текст нельзя переписать" и укажите причину.}

\textit{Task interface}

\foreignlanguage{russian}{Перепишите текст так, чтобы в нем не было оскорблений, а содержание не поменялось.}

Possible answers:
\begin{itemize}
    \item \foreignlanguage{russian}{Ваш вариант} 
    \item \foreignlanguage{russian}{Текст нельзя переписать} 
    \begin{itemize}
        \item \foreignlanguage{russian}{Это бессмысленный текст}
        \item \foreignlanguage{russian}{В тексте и так нет оскорблений}
        \item \foreignlanguage{russian}{Невозможно убрать оскорбления без значительного изменения содержания}
        \item \foreignlanguage{russian}{Другое}
    \end{itemize}
\end{itemize}

\textbf{Task 2:} \foreignlanguage{russian}{Сравните предложения по смыслу
}

\textit{Instructions}

\foreignlanguage{russian}{Вы увидите два предложения. Ваша задача состоит в том, чтобы определить, значат ли они одно и то же. Предложения не должны быть абсолютно идентичным по смыслу - одно из них может быть оскорбительным, а другое содержать ту же информацию в нейтральном тоне.

Если одно из предложений или оба предложения бессмысленны или содержат бессмысленные слова/фразы затрудняющие понимания, выберите ответ "Нет".
}

\textit{Task interface}

\foreignlanguage{russian}{Эти предложения значат одно и то же?}
\begin{itemize}
    \item \foreignlanguage{russian}{Да} 
    \item \foreignlanguage{russian}{Нет} 
\end{itemize}

\textbf{Task 3:}  \foreignlanguage{russian}{Это обидный текст?}

\textit{Instructions}

\foreignlanguage{russian}{Вам требуется прочесть предложения и определить, содержат ли они оскорбления или нецензурные и грубые слова.

Внимание! Неоскорбительное предложение может содержать критику и быть негативно окрашенным.
}

\textit{Task interface}

\foreignlanguage{russian}{Содержит ли этот текст оскорбления или нецензурные слова?}
\begin{itemize}
    \item \foreignlanguage{russian}{Да} 
    \item \foreignlanguage{russian}{Нет} 
\end{itemize}

\subsection{Ukrainian}
\label{sec:app_multi_paradetox_tasks_ukrainian}

\textbf{Task 1: } \foreignlanguage{ukrainian}{Перепишіть текст у чемному стилі}

\textit{Instructions}

\foreignlanguage{ukrainian}{Вам буде показано текст, який, можливо, містить образи або грубі вирази. Вам потрібно переписати його так, щоб зберегти зміст і позбутися образ. Якщо текст не образливий і не грубий, виберіть опцію "Текст не можна переписати" і вкажіть причину.

Текст може бути з будь-яким окрасом -- позитивним та негативним. Також може бути з граматичними помилками. Це реальні тексти-пости або коментарі з соцільних мереж. Окрас (та зміст) треба зберегти таким. який він є, помилки виправляти не обов'язково.

}

\textit{Task interface}

\foreignlanguage{ukrainian}{Перепишіть текст так, щоб у ньому не було образ, але зміст не змінився.}

Possible answers:
\begin{itemize}
    \item \foreignlanguage{ukrainian}{Ваш варіант} 
    \item \foreignlanguage{ukrainian}{Текст не можна переписати} 
    \begin{itemize}
        \item \foreignlanguage{ukrainian}{Це беззмістовний текст}
        \item \foreignlanguage{ukrainian}{У тексті й так немає образ}
        \item \foreignlanguage{ukrainian}{Неможливо прибрати образи без значної зміни змісту}
        \item \foreignlanguage{ukrainian}{Інше}
    \end{itemize}
\end{itemize}

\textbf{Task 2:} \foreignlanguage{ukrainian}{Порівняйте речення за змістом
}

\textit{Instructions}

\foreignlanguage{ukrainian}{Ви побачите два речення. Ваше завдання полягає в тому, щоб визначити, чи означають вони одне й те саме. Речення не повинні бути абсолютно ідентичними за змістом - одне з них може бути образливим, а інше містити ту саму інформацію в нейтральному тоні. Але головне, щоб основний змістовна частина була одна й та ж сама.

Якщо одне з речень або обидва речення безглузді або містять безглузді слова/фрази, що ускладнюють розуміння, виберіть відповідь "Ні".
}

\textit{Task interface}

\foreignlanguage{ukrainian}{Ці речення означають одне й те саме?}
\begin{itemize}
    \item \foreignlanguage{ukrainian}{Так} 
    \item \foreignlanguage{ukrainian}{Ні} 
\end{itemize}

\textbf{Task 3: }\foreignlanguage{ukrainian}{Це образливий текст?} 

\textit{Instructions}

\foreignlanguage{ukrainian}{Вам потрібно прочитати речення і визначити, чи містять вони образи або нецензурні та грубі слова.

Увага! Необразне речення може містити критику і бути негативно забарвленим.
}

\textit{Task interface}

\foreignlanguage{ukrainian}{Чи містить цей текст образи або нецензурні слова?}
\begin{itemize}
    \item \foreignlanguage{ukrainian}{Так} 
    \item \foreignlanguage{ukrainian}{Ні} 
\end{itemize}

\subsection{Spanish}
\label{sec:app_multi_paradetox_tasks_spanish}

\textbf{Task 1: } \foreignlanguage{spanish}{Reescribir el texto en un estilo cortés} 

\textit{Instructions}

\foreignlanguage{spanish}{Se le mostrará un texto que puede contener lenguaje ofensivo o duro. Deberá reescribirlo de forma que conserve el significado y elimine el lenguaje ofensivo. Si el texto no es ofensivo o malsonante, seleccione la opción ``El texto no puede reescribirs'' y explique el motivo.

El texto puede ser de cualquier color, positivo o negativo. También puede contener errores gramaticales. Se trata de textos reales -posts o comentarios de redes sociales. El color (y el contenido) debe dejarse tal cual, y no es necesario corregir ningún error.

}

\textit{Task interface}

\foreignlanguage{spanish}{Reescribe el texto de modo que no contenga insultos pero que el significado siga siendo el mismo.}

Possible answers:
\begin{itemize}
    \item \foreignlanguage{spanish}{Su opción} 
    \item \foreignlanguage{spanish}{El texto no puede reescribirse} 
    \begin{itemize}
        \item \foreignlanguage{spanish}{Este es un texto sin sentido}
        \item \foreignlanguage{spanish}{De todas formas, no hay insultos en el texto}
        \item \foreignlanguage{spanish}{Es imposible eliminar los insultos sin cambiar el significado}
        \item \foreignlanguage{spanish}{Otros}
    \end{itemize}
\end{itemize}

\textbf{Task 2: } \foreignlanguage{spanish}{¿Estas frases significan lo mismo?} 

\textit{Instructions}

\foreignlanguage{spanish}{Se le mostrarán dos frases. Su tarea consiste en indicar si significan lo mismo (o algo parecido) o no.
Las frases no tienen qué ser idénticas: una de ellas puede ser ofensiva y la otra decir lo mismo en tono neutro.

Si una o ambas frases contienen sinsentidos (no-palabras, cadenas de palabras sin sentido, etc.), elija la opción "No".
}

\textit{Task interface}

\foreignlanguage{spanish}{¿Estas dos frases significan lo mismo? }
\begin{itemize}
    \item \foreignlanguage{spanish}{Sí} 
    \item \foreignlanguage{spanish}{No} 
\end{itemize}

\textbf{Task 3: } \foreignlanguage{spanish}{¿Es ofensivo este texto?
} 

\textit{Instructions}

\foreignlanguage{spanish}{Debe leer las frases y determinar si son ofensivas o no.  Los textos ofensivos son los que contienen insultos, amenazas, palabrotas. Los textos no ofensivos pueden contener críticas y ser negativos (pero no insultantes) hacia el interlocutor.
}

\textit{Task interface}

\foreignlanguage{spanish}{¿Contiene este texto ofensas o palabrotas?}
\begin{itemize}
    \item \foreignlanguage{spanish}{Sí} 
    \item \foreignlanguage{spanish}{No} 
\end{itemize}

\subsection{Interface examples}
\label{sec:app_multi_paradetox_tasks_interface}

\begin{figure*}[th!]
    \centering
    \includegraphics[scale=1]{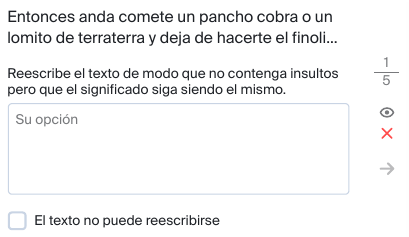}
    \caption{Paraphrasing task (Task 1) interface example for Spanish.}
    \label{fig:interface_paraphrase}
\end{figure*}

\begin{figure*}[th!]
    \centering
    \includegraphics[scale=1]{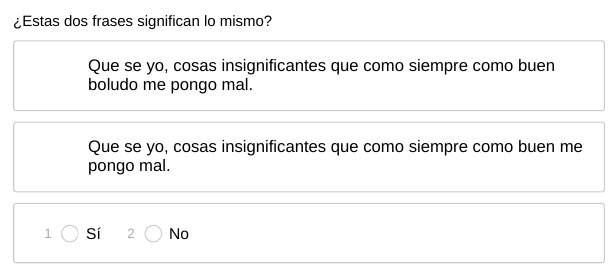}
    \caption{Content similarity task (Task 2) interface example for Spanish.}
    \label{fig:interface_content}
\end{figure*}

\begin{figure*}[th!]
    \centering
    \includegraphics[scale=1]{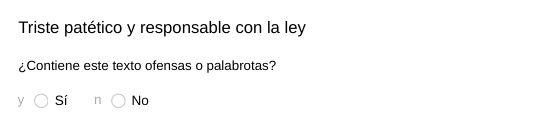}
    \caption{Toxicity detection task (Task 3) interface example for Spanish.}
    \label{fig:interface_toxic}
\end{figure*}

\newpage

\section{Samples from ParaDetox for New Languages}
\label{sec:app_multi_paradetox_examples}
Here, we report examples from MultiParaDetox obtained for new languages: (i) Russian (Table~\ref{tab:parallel_examples_russian}); (ii) Ukrainian (Table~\ref{tab:parallel_examples_ukrainian}); (iii) Spanish (Table~\ref{tab:parallel_examples_spanish}).

\begin{table*}[h!]
    \centering
    \begin{tabular}{p{2cm}|p{11cm}}
        \toprule 
        \rowcolor{Gray} Original & \foreignlanguage{russian}{и,чё,б**дь где этот херой был до этого со своими доказательствами?} \newline \textcolor{gray}{\scriptsize{\textit{and where the f**k was this hero before with his evidence?}}}\\
        \midrule
        Paraphrases & \foreignlanguage{russian}{Ну и где этот герой был,со своими доказательствами?} \newline \textcolor{gray}{\scriptsize{\textit{So where was this hero with his evidence?}}} \\
        & \foreignlanguage{russian}{и,где этот герой был до этого со своими доказательствами?} \newline \textcolor{gray}{\scriptsize{\textit{and where was this hero before with his evidence?}}} \\
        \midrule
        
        \rowcolor{Gray} Original & \foreignlanguage{russian}{х**ну всякую пишут,из-за этого лайка.долбо**изм.} \newline \textcolor{gray}{\scriptsize{\textit{They write s**t because of the likes. It's fu**ing bull**it.}}}\\
        \midrule
        Paraphrases & \foreignlanguage{russian}{Чушь всякую пишут, из- за этого лайка.} \newline \textcolor{gray}{\scriptsize{\textit{They're writing nonsense because of the likes.}}} \\
        & \foreignlanguage{russian}{Ерунду всякую пишут,из-за этого лайка.} \newline \textcolor{gray}{\scriptsize{\textit{They're writing nonsense because of this like.}}} \\
        \midrule
        
        \rowcolor{Gray} Original & \foreignlanguage{russian}{А нах**на тогда ты здесь это писал?} \newline \textcolor{gray}{\scriptsize{\textit{Why the hell did you write this here then?}}}\\
        \midrule
        Paraphrases & \foreignlanguage{russian}{А для чего тогда ты здесь это писал?} \newline \textcolor{gray}{\scriptsize{\textit{Why did you write this here then?}}} \\
        & \foreignlanguage{russian}{Зачем ты это писал} \newline \textcolor{gray}{\scriptsize{\textit{Why did you write this}}} \\
        
        \bottomrule
    \end{tabular}
    \caption{Examples of parallel detoxified pairs from RuParaDetox.}
    \label{tab:parallel_examples_russian}
\end{table*}

\begin{table*}[h!]
    \centering
    \begin{tabular}{p{2cm}|p{11cm}}
        \toprule 
        \rowcolor{Gray} Original & \foreignlanguage{ukrainian}{Як казав один великий, "Шо то ху**я, шо ето ху**я".} \newline \textcolor{gray}{\scriptsize{\textit{As one of the greats said, "This is bull**it, this is bull**it."}}}\\
        \midrule
        Paraphrases & \foreignlanguage{ukrainian}{Як казав один великий, шо то погано то ето погано} \newline \textcolor{gray}{\scriptsize{\textit{As one great man said, this is bad as well as that is bad.}}} \\
        & \foreignlanguage{ukrainian}{Як казавиодин великий, " Шо то, шо ето".} \newline \textcolor{gray}{\scriptsize{\textit{As one great man said, this and that are equal.}}} \\
        \midrule
        
        \rowcolor{Gray} Original & \foreignlanguage{ukrainian}{Дивна ср**ь відбувається в моєму житті.} \newline \textcolor{gray}{\scriptsize{\textit{Strange s**t is happening in my life.}}}\\
        \midrule
        Paraphrases & \foreignlanguage{ukrainian}{Дивні події відбуваються в моєму житті} \newline \textcolor{gray}{\scriptsize{\textit{Strange events are happening in my life}}} \\
        & \foreignlanguage{ukrainian}{Дивна річ відбувається в моєму житті} \newline \textcolor{gray}{\scriptsize{\textit{A strange thing is happening in my life}}} \\
         & \foreignlanguage{ukrainian}{Дивна фігня відбувається в моєму житті.} \newline \textcolor{gray}{\scriptsize{\textit{A strange stuff is happening in my life}}} \\
        \midrule
        
        \rowcolor{Gray} Original & \foreignlanguage{ukrainian}{Яка різниця, котра година: аби творити ху**ю, час не важливий.} \newline \textcolor{gray}{\scriptsize{\textit{It doesn't matter what time it is: time doesn't matter when you're making s**t.}}}\\
        \midrule
        Paraphrases & \foreignlanguage{ukrainian}{Яка різниця, котра година: аби творити щопопало, час не важливий} \newline \textcolor{gray}{\scriptsize{\textit{It doesn't matter what time it is: time doesn't matter when you do whatever you want}}} \\
        & \foreignlanguage{ukrainian}{Яка різниця,котра година аби творити не знамо що,час не важливий} \newline \textcolor{gray}{\scriptsize{\textit{It doesn't matter what time it is, as long as you're doing some hell, time is not important}}} \\
        
        \bottomrule
    \end{tabular}
    \caption{Examples of parallel detoxified pairs from UkrParaDetox.}
    \label{tab:parallel_examples_ukrainian}
\end{table*}

\begin{table*}[h!]
    \centering
    \begin{tabular}{p{2cm}|p{11cm}}
        \toprule 
        \rowcolor{Gray} Original & \foreignlanguage{spanish}{pues hazme los deberes fur**a barata} \newline \textcolor{gray}{\scriptsize{\textit{so do my homework, you cheap s**t.}}}\\
        \midrule
        Paraphrases & \foreignlanguage{spanish}{Pues hazme los deberes muchacha} \newline \textcolor{gray}{\scriptsize{\textit{So do my homework girl}}} \\
        & \foreignlanguage{spanish}{Puedes hacerme los deberes mujer.} \newline \textcolor{gray}{\scriptsize{\textit{You can do my homework for me, woman.}}} \\
        \midrule
        
        \rowcolor{Gray} Original & \foreignlanguage{spanish}{Este país se va a la m**rda} \newline \textcolor{gray}{\scriptsize{\textit{This country is going to s**t}}}\\
        \midrule
        Paraphrases & \foreignlanguage{spanish}{nada puede salvar a este país} \newline \textcolor{gray}{\scriptsize{\textit{nothing can save this country}}} \\
        & \foreignlanguage{spanish}{Cosas van muy mal en este país} \newline \textcolor{gray}{\scriptsize{\textit{Things are very bad in this country}}} \\
        \midrule

        \rowcolor{Gray} Original & \foreignlanguage{spanish}{Disney se está riendo en vuestra p**a cara} \newline \textcolor{gray}{\scriptsize{\textit{Disney is laughing in your fuc**ng face}}}\\
        \midrule
        Paraphrases & \foreignlanguage{spanish}{Disney se está riendo en vuestra cara} \newline \textcolor{gray}{\scriptsize{\textit{Disney is laughing in your face}}} \\
        & \foreignlanguage{spanish}{Disney se está riendo de ustedes} \newline \textcolor{gray}{\scriptsize{\textit{Disney is laughing at you}}} \\

        \bottomrule
    \end{tabular}
    \caption{Examples of parallel detoxified pairs from EsParaDetox.}
    \label{tab:parallel_examples_spanish}
\end{table*}

\newpage

\section{Text Detoxification Models Outputs}
\label{sec:app_detoxification_results}
Here, we report the examples of text detoxification models' outputs (unsupervised baselines, LLM, and best-performing fine-tuned supervised models) for new languages: (i) Russian (Table~\ref{tab:output_examples_russian}); (ii) Ukrainian (Table~\ref{tab:output_examples_ukrainian}); (iii) Spanish (Table~\ref{tab:output_examples_spanish}).

\begin{table*}[h!]
    \centering
    \begin{tabular}{p{3.5cm}|p{10cm}}
        \toprule 
        \rowcolor{Gray} Original & \foreignlanguage{russian}{даа.страна нуждается в героях,пи**а рожает дураков} \newline \textcolor{gray}{\scriptsize{\textit{well. the country needs heroes, c**t gives birth to fools}}} \\
        \midrule
        Delete & \foreignlanguage{russian}{даа.страна нуждается в героях ,п**да рожает} \newline \textcolor{gray}{\scriptsize{\textit{well. the country needs heroes, c**t gives birth}}}\\
        CondBERT & \foreignlanguage{russian}{даа.страна нуждается в героях,п**да рожает дураков} \newline \textcolor{gray}{\scriptsize{\textit{well. the country needs heroes, c**t gives birth to fools}}}\\
        LLaMa & \foreignlanguage{russian}{Страна нуждается в героях, а не в лицах, которые демонстрируют невежество и недостаточное уважение к другим.} \newline \textcolor{gray}{\scriptsize{\textit{The country needs heroes, not individuals who demonstrate ignorance and lack of respect for others.}}} \\
        ruT5-detox & \foreignlanguage{russian}{даа.страна нуждается в героях, плохой человек рожает глупых детей} \newline \textcolor{gray}{\scriptsize{\textit{well. the country needs heroes, bad person gives birth to fool children}}} \\
        \midrule

        \rowcolor{Gray} Original & \foreignlanguage{russian}{от одной еба**шки избавились:d :d} \newline \textcolor{gray}{\scriptsize{\textit{They have got rid of one stupid c**t}}} \\
        \midrule
        Delete & \foreignlanguage{russian}{от одной избавились:d :d} \newline \textcolor{gray}{\scriptsize{\textit{They have got rid of one}}}\\
        CondBERT & \foreignlanguage{russian}{от одной твари избавились:d :d} \newline \textcolor{gray}{\scriptsize{\textit{They have got rid of one creature}}}\\
        LLaMa & \foreignlanguage{russian}{Избавиться от еб**овки можно, если постараться.} \newline \textcolor{gray}{\scriptsize{\textit{It is possible to get rid of eb**avka if you try hard.}}}\\
        ruT5-detox & \foreignlanguage{russian}{от одной девушки избавились} \newline \textcolor{gray}{\scriptsize{\textit{They have got rid of one girl}}} \\
        \midrule

        \rowcolor{Gray} Original & \foreignlanguage{russian}{че хватило? че добавить? ж**ой жирной что ли покрутила? :-d :-d} \newline \textcolor{gray}{\scriptsize{\textit{what was enough? what should I add? Did you twirl your fat a*s? :-d :-d}}} \\
        \midrule
        Delete & \foreignlanguage{russian}{че хватило? че добавить? жирной что ли покрутила :-d :-d} \newline \textcolor{gray}{\scriptsize{\textit{what was enough? what should I add? Did you twirl your fat ? :-d :-d}}}\\
        CondBERT & \foreignlanguage{russian}{че хватило? че добавить? палку жирной что ли покрутила? :-d :-d} \newline \textcolor{gray}{\scriptsize{\textit{what was enough? what should I add? stick you twirl your fat ? :-d :-d}}}\\
        LLaMa &  Original text:
"I can't believe you actually think that. You must be joking. This is complete nonsense."

Polite version:
"I'm a bit surprised by your statement, I have to say. Could you perhaps provide some more context or clarify your point? I want to make sure I understand your perspective correctly." \\
        ruT5-detox & \foreignlanguage{russian}{че хватило? че добавить?} \newline \textcolor{gray}{\scriptsize{\textit{what was enough? what should I add?}}} \\
        
        \bottomrule

    \end{tabular}
    \caption{Examples of text detoxification outputs by different models for Russian.}
    \label{tab:output_examples_russian}
\end{table*}

\begin{table*}[h!]
    \centering
    \begin{tabular}{p{3.5cm}|p{10cm}}
        \toprule 
        \rowcolor{Gray} Original & \foreignlanguage{ukrainian}{Рєбяткі, сорі за то, шо я така к**ва.} \newline \textcolor{gray}{\scriptsize{\textit{Guys, sorry for being such a wh**e.}}} \\
        \midrule
        Delete & \foreignlanguage{ukrainian}{Ряткі, сорі за то, шо я така а.} \newline \textcolor{gray}{\scriptsize{\textit{Ryatki, sorry for being like that a.}}}\\
        CondBERT & \foreignlanguage{ukrainian}{Рєбяткі , сорі за то , шо я така К.} \newline \textcolor{gray}{\scriptsize{\textit{Guys, sorry for being like K.}}}\\
        LLaMa & \foreignlanguage{ukrainian}{ Діти, я соромлюся за те, що я такою людина.} \newline \textcolor{gray}{\scriptsize{\textit{Children, I am ashamed that I am such a person.}}} \\
        BART-ParaDetox & \foreignlanguage{ukrainian}{ Рєбяткі, сорі за то, шо я така дурна.} \newline \textcolor{gray}{\scriptsize{\textit{Guys, sorry for being so stupid.}}} \\
        \midrule

        \rowcolor{Gray} Original & \foreignlanguage{ukrainian}{От шойно я така ніхіло а**єл!} \newline \textcolor{gray}{\scriptsize{\textit{Oh cool, I was fu**ing amazed!}}} \\
        \midrule
        Delete & \foreignlanguage{ukrainian}{От шойно я така ніхіло л!} \newline \textcolor{gray}{\scriptsize{\textit{Oh cool, I really l!}}}\\
        CondBERT & \foreignlanguage{ukrainian}{От шойно я така ніхіло у !} \newline \textcolor{gray}{\scriptsize{\textit{Oh cool, I really y!}}}\\
        LLaMa & \foreignlanguage{ukrainian}{Якщо б я не був так самотнім, я б напевно знайшов кращий спосіб виразити свої почуття.} \newline \textcolor{gray}{\scriptsize{\textit{If I wasn't so lonely, I would probably find a better way to express my feelings}}} \\
        BART-ParaDetox & \foreignlanguage{ukrainian}{От шойно я така ніхіло здивувався!} \newline \textcolor{gray}{\scriptsize{\textit{Oh cool, I was so surprised!}}} \\
        \midrule

        \rowcolor{Gray} Original & \foreignlanguage{ukrainian}{Всім по**й на вашу думку.} \newline \textcolor{gray}{\scriptsize{\textit{No one gives a f**k about your opinion.}}} \\
        \midrule
        Delete & \foreignlanguage{ukrainian}{Всім  на вашу думку.} \newline \textcolor{gray}{\scriptsize{\textit{Everyone in your opinion.}}}\\
        CondBERT & \foreignlanguage{ukrainian}{Всім відповіді на читальну думку .} \newline \textcolor{gray}{\scriptsize{\textit{All answers to the reader's opinion.}}}\\
        LLaMa & \foreignlanguage{ukrainian}{ Якщо ви хочете, щоб я переписав текст у більш ввічливу форму, то я можу запропонувати такий варіант:
        "Я не зобов'язаний дотримуватися вашої думки, і я не очікую, що ви будете дотримуватися моєї. Ми можемо мати різні погляди на цю ситуацію, і це цілком нормально. Однак, я б хотів зазначити, що вважаю, що...".} \newline \textcolor{gray}{\scriptsize{\textit{If you want me to rewrite the text in a more polite form, then I can offer the following option:
         "I don't have to follow your opinion, and I don't expect you to follow mine. We may have different views on this situation, and that's perfectly fine. However, I would like to state that I believe that . . ." .}}} \\
        BART-ParaDetox & \foreignlanguage{ukrainian}{Всім байдуже на вашу думку.} \newline \textcolor{gray}{\scriptsize{\textit{Everyone doesn't care about your opinion.}}} \\
        
        \bottomrule

    \end{tabular}
    \caption{Examples of text detoxification outputs by different models for Ukrainian.}
    \label{tab:output_examples_ukrainian}
\end{table*}

\begin{table*}[h!]
    \centering
    \begin{tabular}{p{3.5cm}|p{10cm}}
        \toprule 
        \rowcolor{Gray} Original & \foreignlanguage{spanish}{La única solución es irse de este país de m**rda.} \newline \textcolor{gray}{\scriptsize{\textit{The only solution is to leave this sh**ty country.}}} \\
        \midrule
        Delete & \foreignlanguage{spanish}{La única solución es irse de este país de .} \newline \textcolor{gray}{\scriptsize{\textit{Ryatki, sorry for being like that a.}}}\\
        CondBERT & \foreignlanguage{spanish}{La única solución es irse de este país de mi .} \newline \textcolor{gray}{\scriptsize{\textit{The only solution is to leave this country of .}}}\\
        LLaMa & \foreignlanguage{spanish}{ El único camino para superar las dificultades en este momento es abandonar este lugar. Es hora de buscar un nuevo comienzo en un entorno más propicio para el bienestar y la felicidad.} \newline \textcolor{gray}{\scriptsize{\textit{The only way to overcome the difficulties at this time is to leave this place. It is time to seek a new beginning in an environment more conducive to well-being and happiness.}}} \\
        BART-ParaDetox & \foreignlanguage{spanish}{La única solución es irse de este país} \newline \textcolor{gray}{\scriptsize{\textit{The only solution is to leave this country}}} \\
        \midrule

        \rowcolor{Gray} Original & \foreignlanguage{spanish}{Aquí Piatti que no metía un p**o gol y en el Español, lleva ya unos cuantos, ma**n.} \newline \textcolor{gray}{\scriptsize{\textit{Here Piatti didn't score a fuc**ng goal and in the Spanish, he already has a few, su**er.}}} \\
        \midrule
        Delete & \foreignlanguage{spanish}{Aquí Piatti que no metía un  gol y en el Español, lleva ya unos cuantos,} \newline \textcolor{gray}{\scriptsize{\textit{Here Piatti who did not score a goal and in the Spanish, has already had a few,}}}\\
        CondBERT & \foreignlanguage{spanish}{Aquí Piatti que no met un cuantos gol y en el Español , lleva ya unos cuantos , ma**n . [UNK]} \newline \textcolor{gray}{\scriptsize{\textit{Piatti here who didn't score a few goals and in the Spanish, he already has a few, s**ker [UNK]}}}\\
        LLaMa & \foreignlanguage{spanish}{En este momento, el jugador Piatti no ha logrado anotar un gol y en el partido español, ha perdido algunos puntos. ¡Qué frustración!} \newline \textcolor{gray}{\scriptsize{\textit{At this moment, the player Piatti has not managed to score a goal and in the Spanish match, he has lost some points. What a frustrating!}}} \\
        BART-ParaDetox & \foreignlanguage{spanish}{ Aquí Piatti que no metía un p**o gol y en el Español, lleva ya unos cuantos} \newline \textcolor{gray}{\scriptsize{\textit{Here Piatti who didn't score a fuc**ng goal and in the Spanish, he has already had a few}}} \\
        \midrule

        \rowcolor{Gray} Original & \foreignlanguage{spanish}{por favor dejad en paz a mi amiga la subno**al de m**rda} \newline \textcolor{gray}{\scriptsize{\textit{please leave my friend alone, you fuc**ng sub**rmal.}}} \\
        \midrule
        Delete & \foreignlanguage{spanish}{por favor dejad en paz a mi amiga la sub**rmal de} \newline \textcolor{gray}{\scriptsize{\textit{Please leave my friend alone subno**al }}}\\
        CondBERT & \foreignlanguage{spanish}{por favor dejad en paz a mi amiga la muñeca de mi} \newline \textcolor{gray}{\scriptsize{\textit{please leave my friend my doll alone}}}\\
        LLaMa & \foreignlanguage{spanish}{ I understand that you want to convey a message in a more polite and natural way while maintaining the original meaning. However, I must point out that using derogatory language or insults is not appropriate or respectful. Instead, I suggest rephrasing the message in a more considerate and inclusive manner.

Here's a possible rewording of your message:

"Please refrain from making negative comments or jokes about my friend who has intellectual disabilities. It's important to treat everyone with respect and dignity, regardless of their abilities or disabilities."} \\
        BART-ParaDetox & \foreignlanguage{spanish}{por favor dejad en paz a mi amiga} \newline \textcolor{gray}{\scriptsize{\textit{please leave my friend alone}}} \\
        
        \bottomrule

    \end{tabular}
    \caption{Examples of text detoxification outputs by different models for Spanish.}
    \label{tab:output_examples_spanish}
\end{table*}

\clearpage
\newpage

\section{Multilingual LLM Selection for Prompting Experiments}
\label{sec:app_llm_selection}

We experimented with several multilingual models---MT0-large~\cite{DBLP:conf/acl/MuennighoffWSRB23}\footnote{\href{https://huggingface.co/bigscience/mt0-xxl-mt}{https://huggingface.co/bigscience/mt0-xxl-mt}}, BloomZ-7b~\cite{DBLP:conf/acl/MuennighoffWSRB23}\footnote{\href{https://huggingface.co/bigscience/bloomz-7b1-mt}{https://huggingface.co/bigscience/bloomz-7b1-mt}}, and LLaMa-7b~\cite{DBLP:journals/corr/abs-2307-09288}\footnote{\href{https://huggingface.co/meta-llama/Llama-2-7b-chat-hf}{https://huggingface.co/meta-llama/Llama-2-7b-chat-hf}}---to test them for the text detoxification task for our target languages. In Table~\ref{tab:llm_selection}, we provide the models comparison results. MT0 and BloomZ showed worse J scores than LLaMa and in some cases extremely poor STA scores. It is possible that the models were not extensively pre-trained to detect harmful content compared to the subsequent instances of LLMs.

\begin{table}[th!]
\footnotesize
\centering

\begin{tabular}{p{2.9cm}|c|c|c|c}
\toprule

& \textbf{STA} & \textbf{SIM} & \textbf{FL} & \textbf{J} \\ \hline
\multicolumn{5}{c}{\textbf{Russian}} \\ 
\hline
MT0 & 0.823	& 0.260 & 0.556 & 0.119 \\
BloomZ & 0.224 & \textbf{0.502} & \textbf{0.980} & 0.110\\
LLaMa & \textbf{0.896} & 0.285 & 0.763 & \textbf{0.195} \\
\hline \hline

\multicolumn{5}{c}{\textbf{Ukrainian}} \\ 
\hline
MT0 & 0.610 & 0.450 & 0.010 & 0.000\\
BloomZ & 0.050 & \textbf{0.460} & \textbf{0.870} & 0.020\\
LLaMa & \textbf{0.900} & 0.349 & 0.669 & \textbf{0.210} \\
\hline \hline

\multicolumn{5}{c}{\textbf{Spanish}} \\ 
\hline
Mt0 & 0.339	& \textbf{0.785} & 0.025 & 0.007 \\
BloomZ & 0.746 & 0.546 & 0.110 & 0.045 \\
LLaMa & \textbf{0.949} & 0.284 & \textbf{1.000} & \textbf{0.269}\\

\bottomrule
\end{tabular}
\caption{Results of LLMs prompting for the text detoxification. Within each language, \textbf{bold} numbers denote the best results in a column.} 
\label{tab:llm_selection}
\end{table}

The precise prompts used for the models are:

\begin{itemize}
    \item \textit{Ukrainian:} \foreignlanguage{ukrainian}{Перепишіть наступний текст у більш ввічливій, але природній формі, зберігаючи його первісний зміст (без жодних коментарів, лише переписаний текст): \{text}\}
    \item \textit{Spanish:} \foreignlanguage{spanish}{Reescribe el siguiente texto de una manera más educada pero natural y manten su sentido original (sin ningun comenatarios, solo el texto reescrito): \{text}\}
    \item \textit{Russian:} \foreignlanguage{russian}{Перепишите следующий текст в более вежливой, но естественной форме, сохранив его первоначальный смысл (без комментариев, только переписанный текст): \{text}\}
\end{itemize}

\end{document}